\documentclass[final]{cvpr}

\usepackage{times}
\usepackage{epsfig}
\usepackage{graphicx}
\usepackage{amsmath}
\usepackage{amssymb}

\usepackage[pagebackref=true,breaklinks=true,colorlinks,bookmarks=false]{hyperref}

\usepackage{booktabs}
\usepackage{multirow}
\usepackage{graphicx}
\usepackage{amsmath,amssymb,amsthm} 
\usepackage[ruled,vlined]{algorithm2e}

\usepackage{subcaption}
\usepackage{footnote}
\makesavenoteenv{tabular}
\makesavenoteenv{table}
\usepackage{color,soul}
\usepackage{xcolor}

\def\GCN{GCN}

\def\ourmethod{GCN-SI}
\def\decisionmaker{decision maker}
\DeclareMathOperator*{\softmax}{Softmax}
\DeclareMathOperator*{\relu}{ReLU}
\DeclareMathOperator*{\maximize}{maximize}

\def\cvprPaperID{7431} 
\def\confYear{CVPR 2021}

\begin{document}

\title{Semi-Supervised Node Classification by Graph Convolutional Networks and Extracted Side Information}

\author{Mohammad Esmaeili\\
The University of Texas at Dallas \\
Richardson, TX 75083-0688, USA.\\
{\tt\small esmaeili@utdallas.edu}
\and
Aria Nosratinia\\
The University of Texas at Dallas \\
Richardson, TX 75083-0688, USA.\\
{\tt\small aria@utdallas.edu}
}

\maketitle

\begin{abstract}
The nodes of a graph existing in a cluster are more likely to connect to each other than with other nodes in the graph. Then revealing some information about some nodes, the structure of the graph (graph edges) provides this opportunity to know more information about other nodes. From this perspective, this paper revisits the node classification task in a semi-supervised scenario by graph convolutional networks (\GCN{}s). The goal is to benefit from the flow of information that circulates around the revealed node labels. The contribution of this paper is twofold. First, this paper provides a method for extracting side information from a graph realization. Then a new \GCN{} architecture is presented that combines the output of traditional \GCN{} and the extracted side information. Another contribution of this paper is relevant to non-graph observations (independent side information) that exists beside a graph realization in many applications. Indeed, the extracted side information can be replaced by a sequence of side information that is independent of the graph structure. For both cases, the experiments on synthetic and real-world datasets demonstrate that the proposed model achieves a higher prediction accuracy in comparison to the existing state-of-the-art methods for the node classification task.
\end{abstract}

\section{Introduction}
\color{black}
Node classification in graphs is generally an unsupervised learning task which refers to clustering (grouping) nodes with similar features. Revealing the labels for a small proportion of nodes transforms the unsupervised node classification task to a semi-supervised learning problem. 
Semi-supervised node classification on a purely graphical observation (a graph realization) has been investigated in the literature on real-world networks by various methods. 
For a brief survey see Section~\ref{section: related work}.

Under the transductive semi-supervised learning setting, the goal is to predict the labels of unlabeled nodes given the adjacency matrix of a graph, the feature matrix containing a set of features for all nodes, and a few revealed node labels. 
There exist various methods for inferring the unlabeled nodes under this setting such as~\cite{kipf2016semi, velivckovic2017graph}. 
Most of the prominent existing methods use either graph-based regularization, graph embedding, or graph convolutional neural networks in a node domain or a spectral domain. 

The structure of a graph (graph edges) allows the graph convolutional network to use a set of fixed training nodes to predict unlabeled nodes. 
Increasing the number of fixed training nodes improves the accuracy of the predictions. 
But in practice, a few training nodes are available for the training procedure. In this paper, we use the predicted labels and involve them effectively in the training procedure to increase the final prediction accuracy.
For this aim, first a sequence of side information is extracted directly from the graph. Then the proposed architecture combines the extracted side information with conventional \GCN{}. 

On the other hand, in many applications, non-graph observations (independent side information) exist beside a given graph realization and its node feature matrix. 
See~\cite{saad2018community} and references therein for a brief introduction about the effects of side information on the community detection for generative models. 
In practice, the feature matrix is not independent of the graph structure, while the non-graph observations may be independent. 
Combining the feature matrix with non-graph observations is a challenging task especially for the case in which the quality of side information is not obvious for the estimator.
Our proposed architecture is also able to combine the provided independent side information with the graph structure and its feature matrix. 

In this paper, it is shown that the proposed architecture, that combines the predicted labels with either extracted or provided side information, achieves higher accuracy in comparison to the existing state-of-the-art methods. 
To the best of our knowledge, this is the first time that the predicted labels in a node classification task are revisited by a graph convolutional neural network to improve the accuracy. 
In addition, this is the first time that the performance of graph convolutional networks is investigated in the presence of independent non-graph observations (side information).

\section{Related Work}
\label{section: related work}
Graph-based semi-supervised methods are typically classified into explicit and implicit learning methods. 
In this section, we review the related work in both classes while the focus of this paper is mainly on the graph convolutional networks which belongs to the latter.
\subsection{Explicit Graph-Based Learning}
In the graph-based regularization methods, it is assumed that the data samples are located in a low dimensional manifold. 
These methods use a regularizer to combine the low dimensional data with the graph.
In the graph-based regularization methods, the objective function of optimization is a linear combination of a supervised loss function for the labeled nodes and a graph-based regularization term with a hyperparameter.
The hyperparameter makes a trade-off between the supervised loss function and the regularization term. 
Graph Laplacian regularizer is widely used in the literature: a label propagation algorithm based on Gaussian random fields~\cite{zhu2003semi}, a variant of label propagation~\cite{talukdar2009new},
a regularization framework by relying on the local or global consistency~\cite{zhou2004learning}, manifold regularization~\cite{belkin2006manifold}, a uniﬁed optimization framework for smoothing language models on graph structures~\cite{mei2008general}, and deep semi-supervised embedding~\cite{weston2012deep}. 

Besides the graph Laplacian regularization, there exist other methods based on the graph embedding: DeepWalk~\cite{perozzi2014deepwalk} that uses the neighborhood of nodes to learn embeddings, LINE~\cite{tang2015line} and node2vec~\cite{grover2016node2vec} which are two extensions of DeepWalk using a biased and complex random walk algorithm, and Planetoid~\cite{yang2016revisiting} which uses a random walk-based sampling algorithm instead of a graph Laplacian regularizer for acquiring the context information.

\subsection{Implicit Graph-Based Learning}
Graph convolutional neural networks have attracted increasing attention recently, as an implicit graph-based semi-supervised learning method. Several graph convolutional network methods have been proposed in the literature: a diffusion-based convolution method which produces tensors as the inputs for a neural network~\cite{atwood2016diffusion}, a scalable and shallow graph convolutional network which encodes both the graph structure and the node features~\cite{kipf2016semi}, a multi-scale graph convolution~\cite{abu2018n}, an adaptive graph convolutional networks~\cite{li2018adaptive},
graph attention networks~\cite{velivckovic2017graph}, 
a variant of attention-based graph neural network for semi-supervised learning~\cite{thekumparampil2018attention}, and dual graph convolutional networks~\cite{zhuang2018dual}.
\section{Proposed Semi-Supervised Node Classification Architecture}
\label{section: proposed method}
In this section, we start by stating some quick intuitions to clarify how revealing some node labels may help the estimator to classify other nodes. 
Then we analyze our main idea which uses revealed node labels and a sequence of synthetic noisy side information.
We provide a technique for extracting side information directly from the adjacency matrix of a graph to be used instead of the synthetic side information.
The conventional graph convolutional network for semi-supervised problem is explained based on~\cite{kipf2016semi}.
Finally, we put all together and propose a model for the semi-supervised node classification task using the graph convolutional network and extracted side information. 

\subsection{Intuition}
We start by a simple example to illustrate how revealed node labels may help an estimator to predict the labels of unlabeled nodes. 
Assume that in a given graph with $k$ classes, the labels of all nodes are revealed except for two nodes $i$ and $j$. 
The goal is to classify node $i$. 
A Bayesian hypothesis testing problem with $k$ hypotheses is considered.
Let $D_{i}$ be a vector of random variables such that its $l$-th element denotes the number of edges from node $i$ to other nodes with revealed labels in the cluster $l$.
Also, let $D'_{i}$ be a vector such that its $l$-th element denotes the number of edges from node $i$ to other unlabeled nodes (node $j$ in this example) in the cluster $l$. Note that $D'_{i}$ is an unknown random variable because the estimator does not know that node $j$ belongs to which class.
The random variable $H$ takes the values in the set $\{1,\cdots, k\}$. For node $i$, we want to infer the value of $H$ by observing a realization of $D_{i}$. Then we have to select the most likely hypothesis conditioned on $D_{i}$, i.e.,
\begin{align*}
    \maximize_{k} \mathbb{P}(H=k|D_{i}=d_{i}) ,
\end{align*}
which is the Maximum A Posteriori (MAP) estimator. 
Let $A$ denote the adjacency matrix of the graph.
With no prior distribution on $H$, when $A_{ij}=0$, the MAP estimator is reorganized as
\begin{align}
\label{intuition: map 1}
    \maximize_{k} \mathbb{P}(D_{i}=d_{i}| H=k) ,
\end{align}
which can be solved by $k-1$ pairwise comparisons. When $A_{ij}=1$,
\begin{align*}
    \mathbb{P}(H=k | D_{i}=d_{i}) &= \frac{\mathbb{P}(D_{i}=d_{i}, H=k) }{\mathbb{P}(D_{i}=d_{i})} \\
    &= \frac{\sum_{d'_{i} \in S} \mathbb{P}(D_{i}=d_{i}, D'_{i}=d'_{i}, H=k) }{\mathbb{P}(D_{i}=d_{i})} ,
\end{align*}
where $S \triangleq \{s=\{0,1\}^{k}: s^{T}\mathbf{1}=1\}$. Assume there exists no prior distribution on $H$. Then the MAP estimator is reorganized as 
\begin{align}
\label{intuition: map 2}
    \maximize_{k} \sum_{d'_{i} \in S} \mathbb{P}(D_{i}=d_{i}, D'_{i}=d'_{i}| H=k).
\end{align}
In practice, although the number of unlabeled nodes is much more than unknown nodes in this example, but a comparison between~\eqref{intuition: map 1} and~\eqref{intuition: map 2} simply shows that how revealing true node labels reduces the complexity of optimal estimator. 

\subsection{Problem Definition \& Analysis}
The focus of this paper is on the graph-based semi-supervised node classification. 
For a given graph with $n$ nodes, let $A$ denote an $n \times n$ adjacency matrix and $X$ denote an $n \times m$ feature matrix, where $m$ is the number of features for each node. 
Under a transductive learning setting, the goal is to infer unknown labels $Y_{u}$, given the adjacency matrix $A$, the feature matrix $X$, and $L$ revealed labels denoted by $Y_{l}$ (fixed training nodes). 
Without loss of generality, assume the first $L$ nodes of the graph are the revealed labels. Then $Y \triangleq [Y_{l}, Y_{u}]$ denotes the vector of all node labels (labeled and unlabeled nodes). 
On the other hand, assume there exists a genie that gives us a vector of side information $Y_{s}$ with length $n$ such that conditioned on the true labels, $Y_{s}$ is independent of the graph edges. Without loss of generality, it is assumed that the entries of $Y_{s}$ are a noisy version of the true labels. In this paper, we suppose that the feature matrix $X$ depends on the graph structure, conditioned on the true labels.
To infer the unlabeled nodes, MAP estimator for this configuration is
\begin{align*}
    \mathbb{P}(Y|A,X,Y_{s},Y_{l}) &= \frac{\mathbb{P}(A,X,Y_{s}, Y_{l}|Y) \mathbb{P}(Y)}{\mathbb{P}(A,X,Y_{s}, Y_{l})} \\
    &\propto \mathbb{P}(A,X, Y_{l}|Y) \mathbb{P}(Y_{s}|Y),
\end{align*}
where $Y$ is drawn uniformly from the set of labels, i.e., there is no prior distribution on node labels. Then we are interested in the optimal solution of the following maximization:
\begin{align*}
    f \triangleq  \maximize_{Y} \log \mathbb{P}(A,X, Y_{l}|Y) + \log \mathbb{P}(Y_{s}|Y).
\end{align*}
Assume $\hat{Y}$ and $\Tilde{Y}$ are the primal optimal solutions of maximizing $\log \mathbb{P}(A,X, Y_{l}|Y)$ and $\log \mathbb{P}(Y_{s}|Y)$, respectively. Then, 
\begin{align*}
&\log \mathbb{P}(A,X, Y_{l}|\tilde{Y}) + \log \mathbb{P}(Y_{s}|\Tilde{Y}) \leq f , \\
&f \leq \log \mathbb{P}(A,X, Y_{l}|\hat{Y}) + \log \mathbb{P}(Y_{s}|\Tilde{Y}).
\end{align*}
Equivalently 
\begin{align*}
    \log \mathbb{P}(A,X, Y_{l}|\tilde{Y}) \leq f - \log \mathbb{P}(Y_{s}|\Tilde{Y}) \leq \log \mathbb{P}(A,X, Y_{l}|\hat{Y}).
\end{align*}
For squeezing $f - \log \mathbb{P}(Y_{s}|\Tilde{Y})$ from above and below, it suffices to provide a model to make $\hat{Y}$ and $\tilde{Y}$ as close as possible by changing the entries of training labels $Y_{l}$. For this purpose, a simple graph convolutional network gives $\hat{Y}$ and $\log \mathbb{P}(A,X, Y_{l}|\hat{Y})$ is maximized for a certain $Y_{l}$. Then the goal now is to use $Y_{s}$ and $\hat{Y}$ to make changes in the training nodes $Y_{l}$.  

Note that the side information that is considered in the analysis is independent of the graph structure. 
This assumption can be relaxed. Indeed, the side information can be extracted from either the feature matrix or the adjacency matrix of a graph. 

\subsection{Extracting Side Information}
For extracting side information that is as much as possible independent from the output of the  graph convolutional neural network, the side information is extracted either from the given feature matrix or the adjacency matrix. 
Define the $r$-neighborhood matrix $A_r$ as
\begin{align*}
    [A_{r}]_{ij} \triangleq  \frac{|N_{i}(r) \cap N_{j}(r)|}{|N_{i}(r) \cup N_{j}(r)|}, 
\end{align*}
where $N_{i}(r)$ is the set of nodes that are in a distance with radius $r$ of node $i$.
This quantity evaluates common neighbors of a pair of nodes with distance $r$ and then normalizes that. 
For extracting side information from the adjacency matrix, a classifier is trained by the $r$-neighborhood matrix and the training nodes, while $r$ is a hyperparameter that must be tuned. A similar idea is represented in~\cite{abbe2015community} in which the authors use a variant of $r$-neighborhood matrices and solve a set of linear equations to theoretically determine whether a pair of nodes are in the same cluster or not. On the other hand, for extracting side information from the feature matrix, a classifier is trained directly by the feature matrix $X$ and the training nodes\footnote{When the feature matrix is equal to the identity matrix, side information is extracted only based on the adjacency matrix.}.

\subsection{Proposed Architecture}
\begin{table*}[ht]
\small
\centering
\begin{tabular}{@{}lccccc@{}}
\toprule
Real-World Dataset & Nodes & Edges & Classes & Features & Training Nodes   \\ \midrule
Cora          & $2708$      & $5429$      & $7$        & $1433$         & $140$  \\
Citeseer      & $3327$      & $4732$      & $6$        & $3703$         & $120$  \\
Pubmed        & $19717$      & $44338$      & $3$        & $500$         & $60$  \\
 \bottomrule
\end{tabular}
\caption{The properties of the real-world datasets for the semi-supervised node classification.}
\label{table: real datasets}
\end{table*} 

\begin{table*}[ht]
\small
\centering
\begin{tabular}{@{}ccccccc@{}}
\toprule
Synthetic Dataset      & Nodes & Classes                     & p & q & Training Nodes \\ \midrule
$k$-SBM & $n = 2000$  & $k \in \{3,4,5\}$ & $5 \times \frac{\log n}{n}$  & $1 \times \frac{\log n}{n}$  & $20 k$      \\ \bottomrule
\end{tabular}
\caption{The properties of the synthetic dataset for the semi-supervised node classification.}
\label{table: SBM}
\end{table*}

\begin{table*}[ht]
\small
\centering
\begin{tabular}{@{}lcccc@{}}
\toprule
\multirow{2}{*}{Hyperparameters} & \multirow{2}{*}{Cora} & \multirow{2}{*}{Citeseer} & \multirow{2}{*}{Pubmed} & \multirow{2}{*}{$k$-SBM} \\
                                  &                       &                           &                         &                          \\ \midrule
$P_{th}$                          & $0.55$                      & $0.80$                          & $0.70$                        & $0.50$                         \\
$F_{th}$                          & $0.99$                      & $0.80$                          & $1.00$                        & $0.50$                         \\
$E_{u}$                           & $50$                      & $80$                          & $80$                        & $150$                         \\
Neurons                 & $128$                      & $128$                          & $64$                        & $16$                         \\
Maximum Epochs                    & $250$                      & $200$                          & $200$                        & $300$                         \\
L2 Regularization Factor          &  $8\times 10^{-5}$                     & $8\times 10^{-5}$                           & $4\times 10^{-4}$                        & $5\times 10^{-5}$                         \\
Learning Rate for Phase 1         & $0.01$                      & $0.01$                          & $0.01$                        & $0.01$                         \\
Learning Rate for Phase 2         & $0.005$                      & $0.05$                          & $0.002$                        & $0.01$                         \\
Correlated Recovery Input(s)                                 & $A_{4}$                      & $X$                          & $A_{1}$                        & $A, A_{1}$                         \\
Correlated Recovery Classifier    & GBC                      & GBC                          & GBC                        & \GCN{}                         \\ \bottomrule
\end{tabular}
\caption{Hyperparameters for the proposed architecture experiments.}
\label{table: hyperparameters}
\end{table*}

\begin{table*}[ht]
\small
\centering
\begin{tabular}{@{}llcccccc@{}}
\toprule
\multirow{2}{*}{Classifier} & \multirow{2}{*}{Input(s)} & \multirow{2}{*}{Cora} & \multirow{2}{*}{Citeseer} & \multirow{2}{*}{Pubmed} & \multicolumn{3}{c}{$k$-SBM} \\ \cmidrule(l){6-8} 
                            &                        &                       &                           &                         & $k=3$   & $k=4$   & $k=5$   \\ \midrule
Neural Network              & $X$                    & $80.3$                      & $56.9$                          & $78.5$                        & $66.9$        & $39.7$         & $25.8$        \\
Neural Network              & $A_{r}$                & $83.4$                      & $74.0$                          &  $79.8$                       & $99.1$        & $94.9$         & $83.4$        \\
Gradient Boosting           & $X$                    &  $83.5$                     & $\mathbf{74.8}$                          &  $79.5$                       &  $33.3$       & $24.8$        & $19.8$        \\
Gradient Boosting           & $A_{r}$                &  $\mathbf{84.7}$                      & $73.0$                          & $\mathbf{81.0}$                        & $99.1$        & $95.1$        & $84.8$        \\
Graph Convolution Network                         & $X, A$                 & $83.1$                      &  $74.2$                         &  $79.5$                       & $93.8$        & $82.4$        & $64.3$        \\
Graph Convolution Network                         & $A_{r}, A$             & $83.4$                      & $73.5$                          & $80.4$                        & $\mathbf{99.3}$        & $\mathbf{96.8}$        & $\mathbf{91.2}$        \\
\bottomrule
\end{tabular}
\caption{Accuracy of \ourmethod{} using various classifiers for the correlated recovery (in percent).}
\label{table: classifier comparisons}
\end{table*}

\begin{figure}[ht]
    \centering
    \includegraphics[scale=0.85]{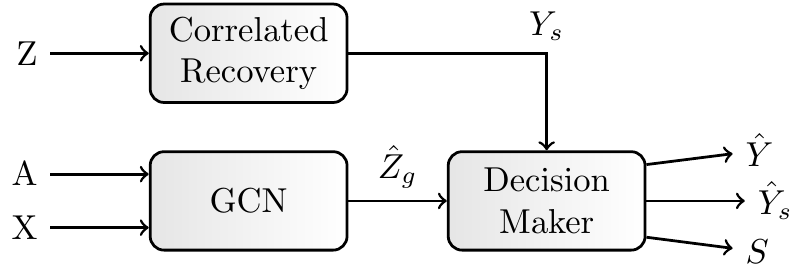}
    \caption{The block diagram of the proposed architecture.}
    \label{fig:model}
\end{figure}

Figure~\ref{fig:model} shows the proposed architecture with three blocks. Throughout this paper, this model is called GCN-SI\footnote{GCN-SI stands on graph convolutional network with side information.} for brevity.  
The \GCN{} block is a conventional graph convolutional network~\cite{kipf2016semi} that takes $X$ and $A$ as inputs.
The graph convolutional network may have several layers. In this paper, we consider a shallow network with two layers. Define $\tilde{A} \triangleq A + \mathbf{I}$ where $\mathbf{I}$ is the identity matrix. Also, define $\tilde{D}$ as a diagonal matrix such that $D_{ii} \triangleq \sum_{j} \tilde{A}_{ij}$.
The graph convolutional network outputs $\hat{Z}_{g}$ as
\begin{align*}
    \hat{Z}_g = f(A,X) = \softmax \left ( \hat{A}~\relu \left( \hat{A}XW^{(0)} \right)W^{(1)}  \right ), 
\end{align*}
where $W^{(0)}$, $W^{(1)}$ are wight matrices that must be trained and $\hat{A} \triangleq \tilde{D}^{-\frac{1}{2}} \tilde{A} \tilde{D}^{-\frac{1}{2}}$. The output is an $n \times k$ matrix, where $k$ is the number of classes.
Note that $\hat{Z}_{g}(i,j)$ determines the probability that the node $i$ belongs to the class $j$ in a graph.

In this paper, side information either is given directly or it should be extracted from the feature or the adjacency matrix.
Then, the correlated recovery block is applied when the side information is not given directly.
The correlated recovery is a simple classifier such as Gradient Boosting.
The input of the correlated recovery block is either the feature matrix $X$ or a variant of the adjacency matrix $A$ (neighborhood matrix $A_r$). The output of the correlated recovery block is $Y_{s}$ which is a vector with length $n$.
In community detection, the correlated recovery refers to the recovering of node labels better than random guessing. 
It is shown that for the case in which the quality of provided side information is poor, the side information still improves the final prediction accuracy in \ourmethod{}.

The \decisionmaker{} decides how to combine the provided side information $Y_{s}$ and the output of the \GCN{} block $\hat{Z}_{g}$. First, the \decisionmaker{} embeds the fixed training labels $Y_{l}$ inside the side information $Y_{s}$, resulting in $\hat{Y}_{s}$ such that $\hat{Y}_{s}(i)=Y(i)$ for all $i \in \{\text{fixed training nodes}\}$. 
The \decisionmaker{} returns $\hat{Y}_s$, the predicted labels $\hat{Y}$ which are vectors with length $n$ and a set of node indices $S$ that are used for defining the loss function. 
Then the loss function for this model is defined as
\begin{align*}
    \mathcal{L}(\hat{Y}_{s}, \hat{Y}, S) \triangleq  \frac{1}{|S|} \sum_{i \in S} \mathcal{L}_{0}(\hat{Y}_{s}(i), \hat{Y}(i)),
\end{align*}
where $\mathcal{L}_{0}(\cdot, \cdot)$ is the cross-entropy loss function, and the index $i$ in $\hat{Y}(i)$ and $\hat{Y}_{s}(i)$ refers to $i$-th entry. 

Let $E$ index the epochs during the training procedure and $E_{u}$ denote the epoch in which the \decisionmaker{} starts to make a change in the training nodes. 
\begin{itemize}
    \item \textbf{Phase (1):} when $E<E_{u}$,  the \decisionmaker{} returns $\hat{Y}$, $\hat{Y}_s$ and $S= \{\text{fixed training nodes\}}$. This phase is similar to a simple graph convolutional network in~\cite{kipf2016semi}.
    
    \item \textbf{Phase (2):} when $E\geq E_{u}$, the \decisionmaker{} uses $\hat{Z}_{g}$ and determines a set of nodes $S_{1}$ such that each element of $S_{1}$ belongs to a specific class with a probability at least $P_{th}$. Note that $P_{th}$ is a threshold that evaluates the quality of the selected nodes. On the other hand, the \decisionmaker{} obtains a set of nodes $S_{2}$ such that for each element of $S_{2}$ both the corresponding side information and the prediction of the graph convolutional network refer to the same class. Then the \decisionmaker{} obtains
    \begin{equation*}
        S \triangleq (S_1 \cap S_2) \cup \{\text{fixed training node indices}\}.
    \end{equation*}
    Phase (2) continues until the prediction accuracy for the fixed training nodes be grater than $F_{th}$; Otherwise, the training continues based on the last obtained set $S$.
\end{itemize}
Algorithm~\ref{Alg: 1} summarizes the role of \decisionmaker{} during the training procedure. Note that $E_{u}$, $P_{th}$, and $F_{th}$ are three hyperparameters that are tuned. 
\begin{algorithm}[ht]
\SetAlgoLined
\textbf{Inputs:} $\hat{Z}_{g}$, $Y_s$\\
$\hat{Y}_s \gets  Y_{s} $ \\
$\hat{Y}_s(i) \gets Y(i)$ for $i\in$ \{Fixed training nodes\} \\
$\hat{Y} \gets \arg\max  \hat{Z}_{g}$ \\
$F \gets \text{Accuracy of fixed training nodes}$ \\
\If{$E<E_u$}{
        $S \gets \{\text{Fixed training nodes}\}$ 
}
\If{$E\geq E_u$ }{
    \eIf{$F \geq F_{th}$}{
    Obtain $S_1$ and $S_2$ based on $P_{th}$\\
    $S \gets (S_1 \cap S_2) \cup \{\text{fixed training nodes}\} $ \\
    Save $S$
    }
    {
    $S \gets$ Load $S$ 
    }
}
\textbf{return} $\hat{Y}_s$, $\hat{Y}$, $S$ 
\caption{Decision Maker}
 \label{Alg: 1}
\end{algorithm}
Assume at epoch $E_{u}$, the optimal solution for maximizing $\mathbb{P}(A,X,Y_{l}|Y)$ is $\hat{Y}_{g}^{E_{u}}$ which is extracted from $\hat{Z}_{g}$. The \decisionmaker{} uses $\hat{Z}_{g}$ and $Y_{s}$ to obtain a set of nodes $S$ that is used for the next training iteration. 
Also, since neural networks are robust to the noisy labels~\cite{rolnick2017deep, hendrycks2018using, ghosh2017robust}, the selected nodes will have enough quality to be involved in the training process by choosing an appropriate value for $P_{th}$. 
Note that the hyperparameter $P_{th}$ determines the quality of the selected nodes.
Then at epoch $E_{u}+1$, the training is based on a new training set $Y_{l}^{E_{u}} \triangleq \{Y_{s}(i): i\in S \}$ which includes the fixed training labels. Let $\hat{Y}_{g}^{E_{u}+1}$ be the optimal solution for maximizing $\mathbb{P}(A,X,Y_{l}^{E_{u}}|Y)$. Note that the side information $Y_{s}$ is more similar to $Y_{l}^{E_{u}}$ than  $Y_{l}$. Then $\tilde{Y}$ is more similar to $\hat{Y}_{g}^{E_{u}+1}$ than $\hat{Y}_{g}^{E_{u}}$ and the main idea follows.

\begin{figure*}
\begin{center}
\begin{subfigure}{0.3\textwidth}
         \centering
         \includegraphics[width=\textwidth]{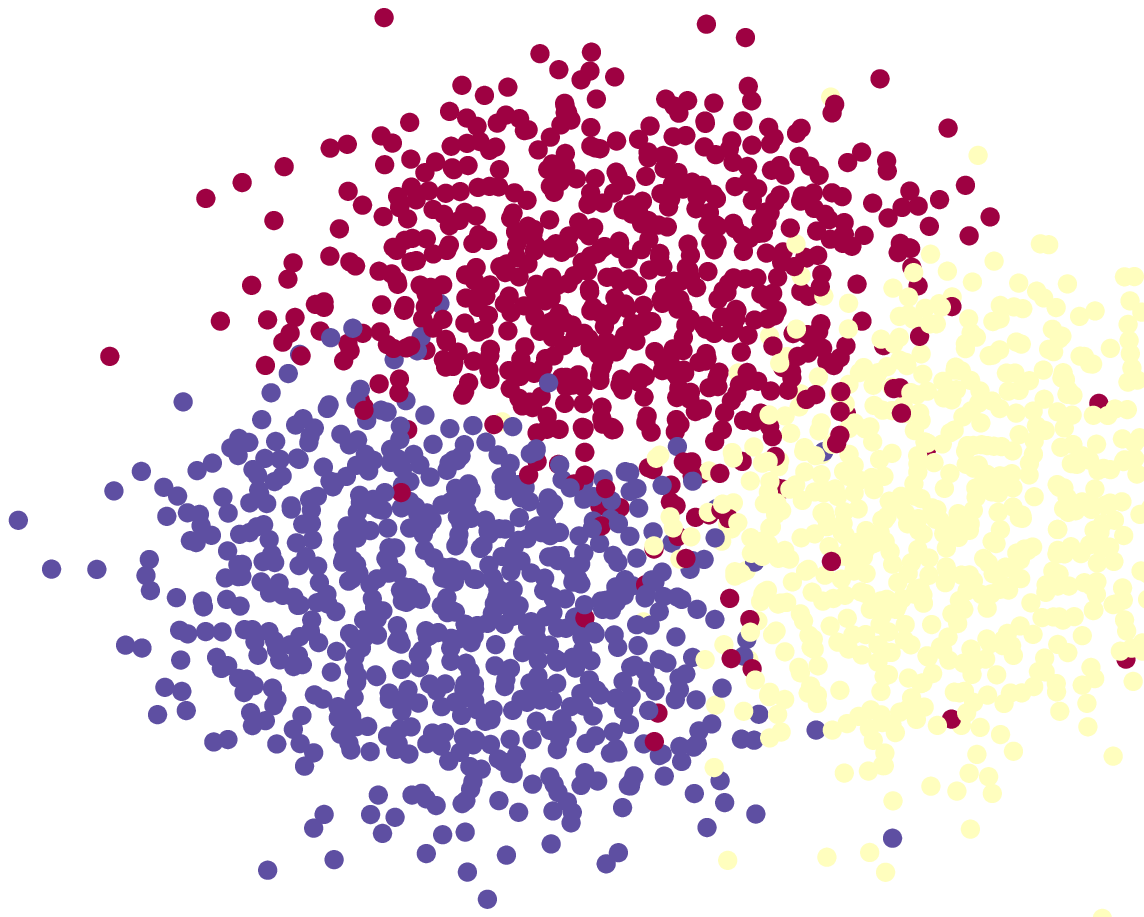}
         \caption{$k=3$}
         \label{fig:k=3}
     \end{subfigure}
     \hfill
     \begin{subfigure}{0.3\textwidth}
         \centering
         \includegraphics[width=\textwidth]{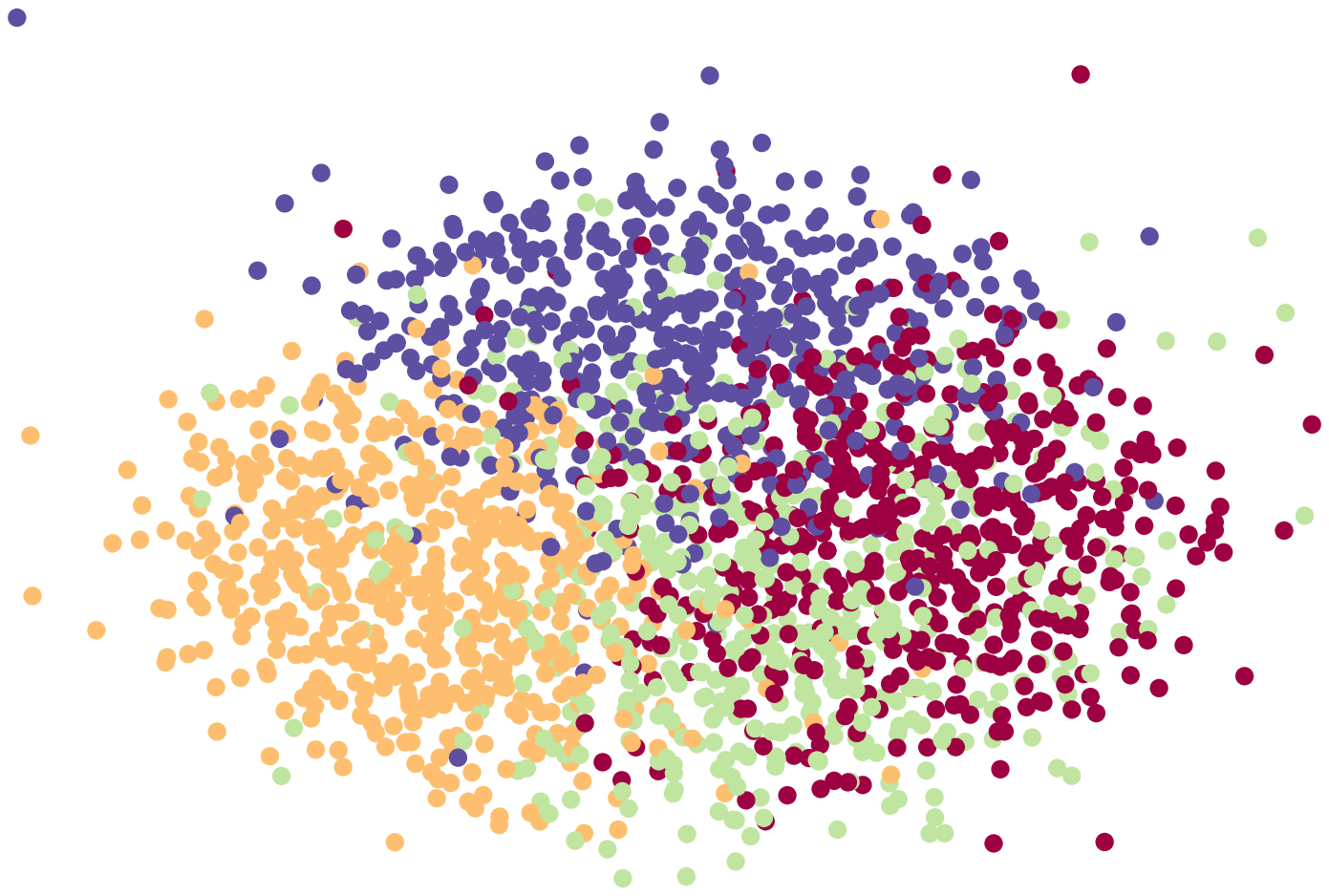}
         \caption{$k=4$}
         \label{fig:k=4}
     \end{subfigure}
     \hfill
     \begin{subfigure}{0.3\textwidth}
         \centering
         \includegraphics[width=\textwidth]{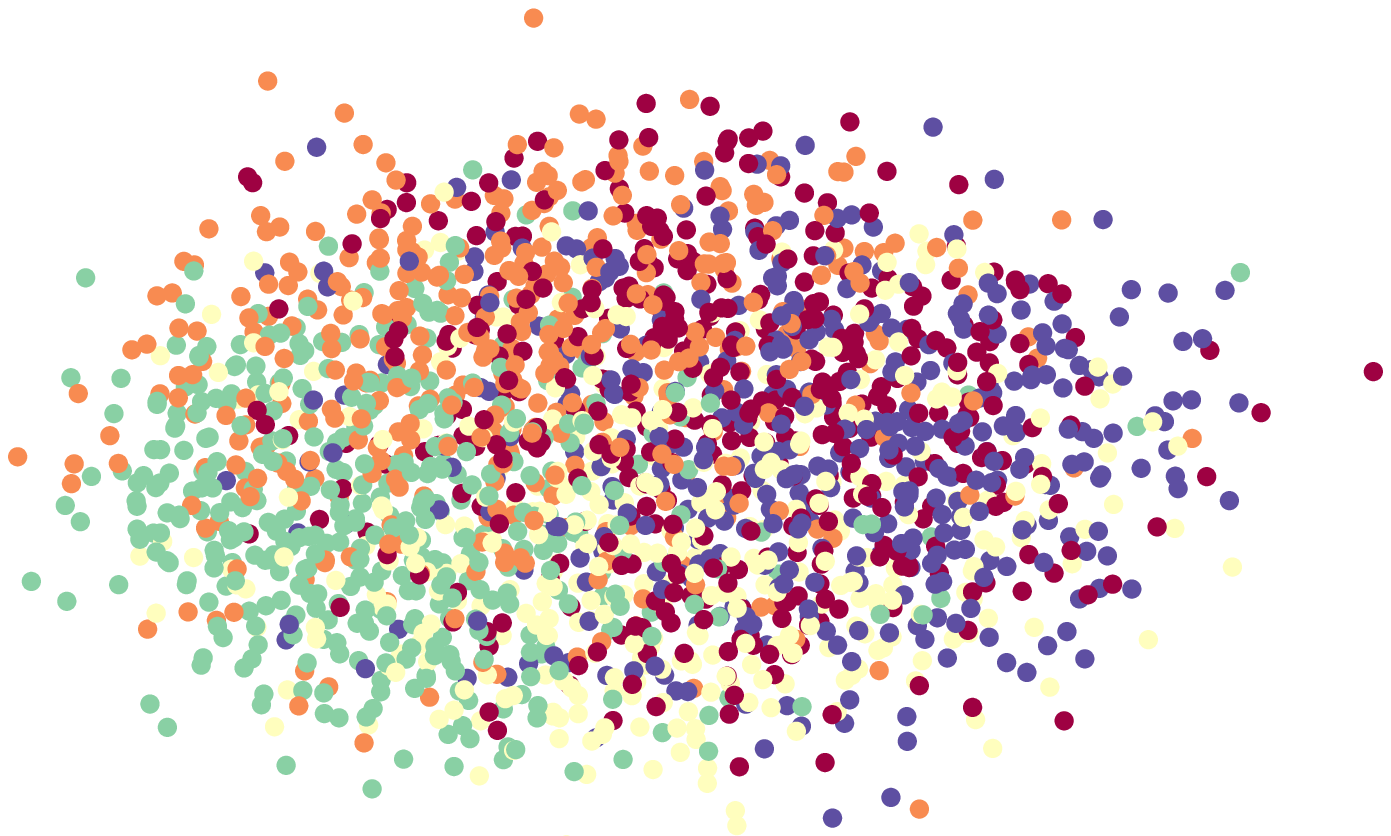}
         \caption{$k=5$}
         \label{fig:k=5}
     \end{subfigure}
\end{center}
\caption{Three realizations of $k$-SBM with $2000$ nodes and different number of classes.}
\label{fig: SBM}
\end{figure*}

\section{Experiments}
The proposed model in Section~\ref{section: proposed method} is tested under a number of experiments on synthetic and real-world datasets: semi-supervised document classification on three real citation networks, semi-supervised node classification under the stochastic block model with different number of classes, and semi-supervised node classification in the presence of noisy labels side information which is independent of graph edges for both the synthetic and real-world datasets. 

\subsection{Datasets \& Side Information}
\textbf{Citation Networks:} Cora, Citeseer, and Pubmed are three common citation networks that have been investigated in previous studies. 
In these networks, articles are considered as nodes. The article citations determine the edges connected to the corresponding node. 
Also, a sparse bag-of-words vector, extracted from the title and the abstract of each article, is used as a vector of features for that node. 
Table~\ref{table: real datasets} shows the properties of these real-world datasets. 

\noindent \textbf{Stochastic Block Model (SBM):} The stochastic block model is a generative model for random graphs which produces graphs containing clusters. Here, we consider a stochastic block model with $n = 2000$ nodes and $k$ classes. Without loss of generality, it is assumed that the true label for each node is drawn uniformly from the set $\{0,\cdots, k-1\}$. Under stochastic block model, if two nodes belong to the same class then an edge is drawn between them with probability $p$; Otherwise, these nodes are connected to each other with probability $q$. 
Table~\ref{table: SBM} summarizes the properties of the stochastic block model in our experiments. Also, Figure~\ref{fig: SBM} shows three realizations of the described generative model with the parameters in Table~\ref{table: SBM}. In this paper, a realization of the stochastic block model, based on the parameters in Table~\ref{table: SBM} with $k$ classes, is briefly called $k$-SBM dataset.

\noindent \textbf{Noisy Labels Side Information:} We consider a noisy version of the true label for each node as synthetic side information. 
This information is given to the \decisionmaker{} to investigate the effect of a non-graph observation which is completely independent of the graph edges. 
Under the noisy labels side information, the \decisionmaker{} observes the true label of each node with probability~$\alpha$; Otherwise, the \decisionmaker{} observes a value that is drawn uniformly from the incorrect labels. 

\subsection{Experimental Settings}
For the \GCN{} block in Figure~\ref{fig:model}, a two-layer graph convolutional network is trained with $\relu$ and $\softmax$ activation functions at the hidden and output layers, respectively. For real-world datasets, we exactly follow the same data splits in~\cite{kipf2016semi} including $20$ nodes per class for training, $500$ nodes for the validation, and $1000$ nodes for the test.   
For $k$-SBM datasets, we follow a data splitting similar to the one used for the real-world datasets. 
Then it is randomly considered $20$ nodes per class for the training, $500$ nodes for the validation, and $1000$ nodes for the test.  
The weights of the neural networks are initialized by the Glorot initialization in~\cite{glorot2010understanding}. Adam~\cite{kingma2014adam} optimizer with specific learning rates for phase (1) and phase (2) is applied. 
Also, the cross-entropy loss is used for all datasets.
Table~\ref{table: hyperparameters} summarizes the values of hyperparameters that are picked for each dataset in the experiments. 

For real-world datasets Gradient Boosting is used as the correlated recovery classifier. Also, for synthetic datasets a graph convolutional network is used to classify the nodes at the correlated recovery block. 

\subsection{Baselines}
For the synthetic datasets either with or without the synthetic side information, the proposed architecture is compared with~\cite{kipf2016semi}. When the synthetic side information is not available, our model benefits from correlated recovery to extract the side information.  
For the real-world datasets, the architecture is compared with several state-of-the-art methods under the transductive semi-supervised learning setting. 
These methods have been listed in Table~\ref{table: state-of-the-art} including graph Laplacian regularized methods~\cite{brandes2007modularity, zhu2003semi, zhou2004learning, yang2016revisiting} and deep graph embedding methods~\cite{velivckovic2017graph, zhuang2018dual, du2017topology, abu2018n}. 
The comparisons are based on the reported prediction accuracy in each paper for each dataset.

\section{Results}

\begin{table*}[ht]
\small
\centering
\begin{tabular}{@{}lccccccc@{}}
\toprule
\multirow{2}{*}{Method} & \multirow{2}{*}{\begin{tabular}[c]{@{}c@{}}Synthetic\\ Side Information\end{tabular}} & \multirow{2}{*}{Cora} & \multirow{2}{*}{Citeseer} & \multirow{2}{*}{Pubmed} & \multicolumn{3}{c}{$k$-SBM} \\ \cmidrule(l){6-8} 
                        &                                                                                       &                       &                           &                         & $k=3$   & $k=4$   & $k=5$   \\ \midrule
\GCN{}~\cite{kipf2016semi}                     & without                                                                               & $81.5$                      & $70.3$                          & $79.0$                        & $96.5$        & $86.9$        & $75.1$        \\
\ourmethod{} (ours)                  & without                                                                               & $84.7$                      & $74.8$                          & $81.0$                        & $99.3$        & $96.7$        & $90.6$        \\
\ourmethod{} (ours)                   & $\alpha=0.7$                                                                          & $85.8$                      &  $75.3$                         & $80.9$                        & $99.4$        & $97.2$        & $93.0$        \\
\ourmethod{} (ours)                  & $\alpha=0.5$                                                                          & $85.1$                      & $74.9$                          & $80.3$                        & $99.2$        & $96.7$        & $90.8$        \\ 
\ourmethod{} (ours)                  & $\alpha=0.3$                                                                          & $84.1$                     & $73.9$                         & $79.4$                       & $99.0$        & $95.5$       & $86.3$        \\
\bottomrule
\end{tabular}
\caption{Accuracy of \ourmethod{} and \GCN{}~\cite{kipf2016semi} in the presence of extracted or synthetic side information (in percent).}
\label{table: side information comparison}
\end{table*}

\begin{table*}[ht]
\small
\centering
\begin{tabular}{@{}lccccccc@{}}
\toprule
\multirow{2}{*}{Method} & \multirow{2}{*}{\begin{tabular}[c]{@{}c@{}}Synthetic\\ Side Information\end{tabular}} & \multirow{2}{*}{Cora} & \multirow{2}{*}{Citeseer} & \multirow{2}{*}{Pubmed} & \multicolumn{3}{c}{$k$-SBM} \\ \cmidrule(l){6-8}     &        &        &    &     & $k=3$   & $k=4$   & $k=5$   \\ \midrule
\GCN{}~\cite{kipf2016semi}      & $\alpha = 0.7$      & $86.4$      & $76.4$      & $82.7$       & $98.4$       & $96.0$      & $92.4$   \\
\ourmethod{} (ours)           & $\alpha = 0.7$      & $88.7$      & $80.6$      & $83.3$       & $98.6$       & $96.2$      & $92.9$   \\
\GCN{}~\cite{kipf2016semi}      & $\alpha = 0.5$      & $83.6$      & $72.6$      & $74.7$       & $87.1$       & $84.2$      & $78.7$   \\
\ourmethod{} (ours)           & $\alpha = 0.5$      & $86.7$      & $77.7$      & $74.5$       & $83.5$       & $83.2$      & $78.6$   \\
\GCN{}~\cite{kipf2016semi}      & $\alpha = 0.3$      & $81.0$      & $68.8$      & $66.8$       & $32.1$       & $42.0$      & $44.4$   \\
\ourmethod{} (ours)           & $\alpha = 0.3$      & $84.0$      & $74.0$      & $60.9$       & $29.1$       & $37.3$      & $43.1$   \\
\bottomrule
\end{tabular}
\caption{Accuracy of \ourmethod{} and \GCN{}~\cite{kipf2016semi} in the presence of synthetic side information embedded in the feature matrix (in percent).}
\label{table: side information with embedding}
\end{table*}

In this section, we report the average prediction accuracy on the test set by running $100$ repeated runs with random initializations for each dataset
\footnote{The code is available online at: \\ \url{https://github.com/mohammadesmaeili/GCNN}}.
The presented results have three parts: investigating the effect of various classifiers (at the correlated recovery block) on the accuracy performance of \ourmethod{}, showing the merit of \ourmethod{} in dealing with a non-graph observation which is independent of the graph edges, and expressing the superiority of \ourmethod{} in comparison to the existing state-of-the-art methods. 
Unless otherwise noted, the experiments in this section follow the hyperparameters represented in Table~\ref{table: hyperparameters}.

Table~\ref{table: classifier comparisons} compares the prediction accuracy of various classifiers at the correlated recovery block in Figure~\ref{fig:model}. 
In Table~\ref{table: classifier comparisons}, for each dataset, either the $r$-neighborhood matrix $A_{r}$ or the feature matrix $X$ is considered as the classifier input.  
For each classifier and each dataset, $A_{r}$ and other classifier hyperparameters have been chosen appropriately to maximize the accuracy on the validation set. 
Note that for $k$-SBM dataset the feature matrix does not exist, i.e., $X=\mathbf{I}$. 
Then the extracted side information only based on the feature matrix is not reliable. 

Table~\ref{table: side information comparison} summarizes the results which compare the proposed architecture with the \GCN{}~\cite{kipf2016semi} for both real-world and synthetic datasets. The results show that without independent side information, the accuracy of \ourmethod{} outperforms the traditional \GCN{} because it benefits from the extracted side information.
Also, Table~\ref{table: side information comparison} makes a comparison between the quality of the extracted side information and the synthetic noisy labels side information with various values for noise parameters $\alpha$. 
The accuracy of the extracted side information in \ourmethod{} for Cora, Citeseer, and Pubmed datasets is around 63\%, 52\%, and 51\% ,respectively.

Note that in Table~\ref{table: side information comparison}, the synthetic side information is not combined with the feature matrix because it is assumed that the quality of the side information is unknown. 
If the synthetic side information has enough and acceptable quality, it can be embedded in the feature matrix. 
This embedding improves the accuracy of both the classical \GCN{} and the proposed architecture. 
But if the side information does not have enough quality, embedding reduces the accuracy of both methods dramatically. 
Considering this fact, Table~\ref{table: side information with embedding} shows the results when the synthetic side information is combined with the feature matrix for both classical \GCN{} and \ourmethod{}. 
Then we need to create a new feature matrix by combining the side information with the feature matrix $X$.
Therefore, for real-world datasets, the new feature matrix is created by stacking the one-hot matrix of synthetic side information to the given feature matrix. 
Also, for synthetic $k$-SBM dataset, the one-hot matrix of side information is used as a newly created feature matrix instead of the identity matrix.

Finally, the accuracy of \ourmethod{} is compared with the reported accuracy of several state-of-the-art methods under the transductive semi-supervised learning setting. 
The results are summarized in Table~\ref{table: state-of-the-art}. 
The proposed architecture achieves higher accuracy in comparison to all existing methods for Cora, Citeseer, and Pubmed datasets. 
The results verify the main idea of this paper for improving the prediction accuracy. Indeed, prediction accuracy increases by revealing more node labels and allowing the nodes of a graph to have more information about the other nodes.  

\begin{figure}[ht]
    \centering
    \includegraphics[scale=0.595]{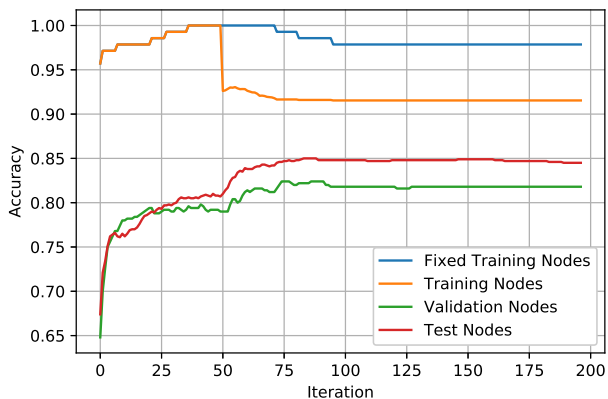}
    \caption{Training on Cora dataset.}
    \label{fig: cora}
\end{figure}

\begin{figure}[ht]
    \centering
    \includegraphics[scale=0.595]{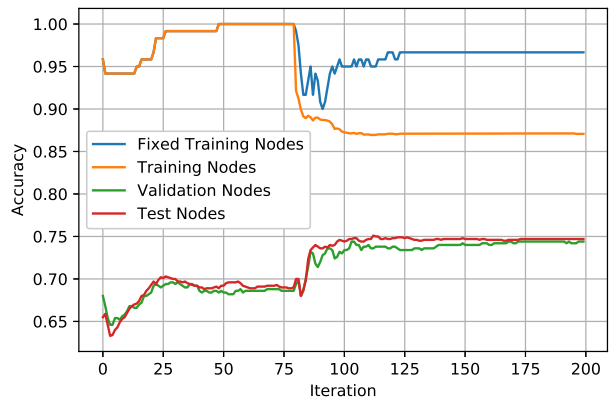}
    \caption{Training on Citeseer dataset.}
    \label{fig: citeseer}
\end{figure}

\begin{figure}[ht]
    \centering
    \includegraphics[scale=0.595]{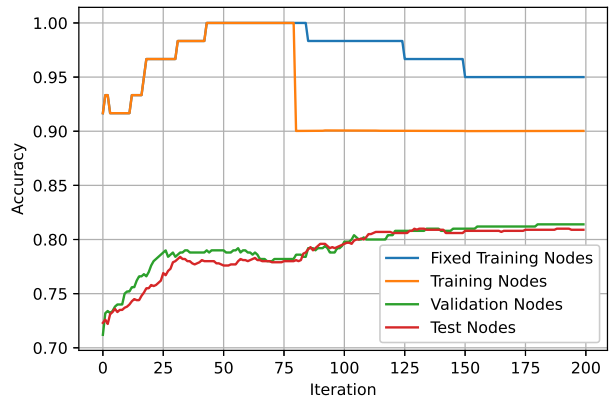}
    \caption{Training on Pubmed dataset.}
    \label{fig: pubmed}
\end{figure}

\begin{figure}[ht]
    \centering
    \includegraphics[scale=0.595]{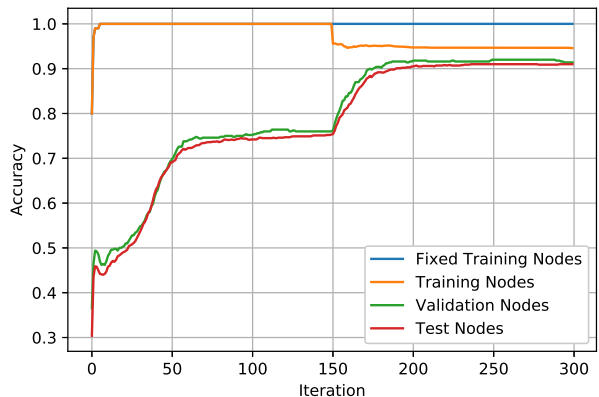}
    \caption{Training on $k$-SBM dataset with $k=5$.}
    \label{fig: k-sbm}
\end{figure}

Figures~\ref{fig: cora}, \ref{fig: citeseer}, \ref{fig: pubmed}, and \ref{fig: k-sbm} are examples of training by \ourmethod{} on Cora, Citeseer, Pubmed, and $k$-SBM with $k=5$, respectively. For each dataset, the accuracy of the test and validation sets show a remarkable difference before and after applying the extracted side information. 

\begin{table}[ht]
\small
\centering
\begin{tabular}{@{}llll@{}}
\toprule
Method & Cora & Citeseer & Pubmed  \\ \midrule 
Modularity Clustering \cite{brandes2007modularity}      & $59.5$   & $60.1$  &  $70.7$  \\
SemiEmb \cite{weston2012deep}      & $59.0$     & $59.6$         & $71.1$     \\
DeepWalk \cite{zhou2004learning}      & $67.2$     & $43.2$         & $65.3$     \\
Gaussian Fields\cite{zhu2003semi}      & $68.0$     & $45.3$         & $63.0$     \\
Planetoid \cite{yang2016revisiting}   & $75.7$   & $64.7$    & $77.2$   \\
DCNN \cite{atwood2016diffusion}      & $76.8$     & -         & $73.0$     \\
\GCN{} \cite{kipf2016semi}      & $81.5$     & $70.3$         & $79.0$     \\
MoNet \cite{monti2017geometric}      & $81.7$     &  -        & $78.8$     \\
N-GCN \cite{abu2018n}      & $83.0$     & $72.2$         & $79.5$     \\
GAT \cite{velivckovic2017graph}      & $83.0$     & $72.5$         & $79.0$     \\
AGNN \cite{thekumparampil2018attention}     & $83.1$     & $71.7$         & $79.9$     \\
TAGCN \cite{du2017topology}     & $83.3$     & $72.5$         & $79.0$     \\
DGCN \cite{zhuang2018dual}     & $83.5$     & $72.6$         & $80.0$ \\ 
LSM-GAT \cite{ma2019flexible}     & $82.9$     & $73.1$         & $77.6$ \\
SBM-GCN \cite{ma2019flexible}     & $82.2$     & $74.5$         & $78.4$ \\ \midrule
\textbf{ \ourmethod{} (ours) }    & \textbf{84.7}     & \textbf{74.8}         & \textbf{81.0}     \\  \bottomrule
\end{tabular}
\caption{Accuracy of various semi-supervised node classification methods (in percent).}
\label{table: state-of-the-art}
\end{table} 



\section{Conclusion \& Future Work}
In this paper, we proposed a new architecture for the semi-supervised node classification task.
In this model, first side information is extracted from the graph structure. For extracting the side information, we introduced a method based on the number of common existing nodes in a certain distance between a pair of nodes.  
Then the extracted side information is combined with a graph convolutional network to estimate the unknown labels using the adjacency matrix, the feature matrix, and the revealed labels. 

Comparisons between the proposed architecture and the existing state-of-the-art methods verified that combining graph convolutional networks and extracted side information results in a higher accuracy performance.
Also, we indicated how the proposed model outperforms the conventional \GCN{}~\cite{kipf2016semi} in the presence of both a graph realization and an independent graph side information. 

The main contribution of this paper is investigated under the \GCN{}~\cite{kipf2016semi} while it remains an open problem to investigate the proposed idea with other semi-supervised classification methods. 
Also, investigating the role of independent graph side information (in a general form consisting of multiple features with finite cardinality) is still an open problem for other state-of-the-art methods. 


{\small
\bibliographystyle{unsrt}
\bibliography{References}
}

\end{document}